\documentclass[11pt]{article} 

\usepackage[utf8]{inputenc} 

\usepackage{geometry} 
\usepackage{graphicx} 

\usepackage{booktabs} 
\usepackage{array} 
\usepackage{paralist} 
\usepackage{verbatim} 
\usepackage{subfig} 
\usepackage{algorithm}
\usepackage{algcompatible}

\usepackage{fancyhdr} 
\usepackage{sectsty}
\allsectionsfont{\sffamily\mdseries\upshape} 

\usepackage[nottoc,notlof,notlot]{tocbibind} 
\usepackage[titles,subfigure]{tocloft} 

\usepackage{algorithm}
\usepackage{algcompatible}
\usepackage[ampersand]{easylist}
\usepackage{algorithm}
\usepackage{pbox}
\usepackage{amsfonts,amssymb}
\usepackage{amsmath}
\usepackage{changepage}
\usepackage{tikz}
\usepackage{soul}
\usepackage{multirow}
\usepackage{booktabs}
\usepackage{threeparttable}
\usepackage{tabularx}
\usepackage{hyperref}
\hypersetup{colorlinks,linkcolor={blue},citecolor={blue},urlcolor={blue}}
\usepackage{url}
\usepackage{pbox}
\usepackage{makecell}
\usepackage{authblk}
\usepackage{notoccite}
\usepackage{bm}
\newsavebox{\tablebox}

\usepackage{array}
\newcolumntype{L}[1]{>{\raggedright\arraybackslash}p{#1}}

\begin{document}

\title{An Approach for Adaptive Automatic Threat Recognition Within 3D Computed Tomography Images for Baggage Security Screening}

\author[1]{Qian Wang \thanks{Corresponding: Qian Wang, Email: qian.wang@durham.ac.uk/qian.wang173@hotmail.com \\ This paper is based upon work supported by the U.S. Department of Homeland security under Award Number 2013-ST-061-E001. The views and conclusions contained in this document are those of the authors and should not be interpreted as necessarily representing the official policies, either expressed or implied, of the U.S. Department of Homeland Security.}}
\author[1,3]{Khalid N. Ismail}
\author[1,2]{Toby P. Breckon}
\affil[1]{Department of Computer Science, Durham University, United Kingdom}
\affil[2]{Department of Engineering, Durham University, United Kingdom}
\affil[3]{Information Technology Department, Faculty of Computers and Information, Menoufia University, Egypt}

\date{} 
\maketitle
\begin{abstract}
\textbf{BACKGROUND:} The screening of baggage using X-ray scanners is now routine in aviation security with automatic threat detection approaches, based on 3D X-ray computed tomography (CT) images, known as Automatic Threat Recognition (ATR) within the aviation security industry. These current strategies use pre-defined threat material signatures in contrast to adaptability  towards new and emerging threat signatures. To address this issue, the concept of adaptive automatic threat recognition (AATR) was proposed in previous work. 

\textbf{OBJECTIVE:} In this paper, we present a solution to  AATR based on such X-ray CT baggage scan imagery. This aims to address the issues of rapidly evolving threat signatures within the screening requirements. Ideally, the detection algorithms deployed within the security scanners should be readily adaptable to different situations with varying requirements of threat characteristics (e.g., threat material, physical properties of objects). 

\textbf{METHODS:} We tackle this issue using a novel adaptive machine learning methodology with our solution consisting of a multi-scale 3D CT image segmentation algorithm, a multi-class support vector machine (SVM) classifier for object material recognition and a strategy to enable the adaptability of our approach. Experiments are conducted on both open and sequestered 3D CT baggage image datasets specifically collected for the AATR study. 

\textbf{RESULTS:} Our proposed approach performs well on both recognition and adaptation. Overall our approach can achieve the probability of detection around 90\% with a probability of false alarm below 20\%. 

\textbf{CONCLUSIONS:} Our AATR shows the capabilities of adapting to varying types of materials, even the unknown materials which are not available in the training data, adapting to varying required probability of detection and adapting to varying scales of the threat object.
\end{abstract}
\providecommand{\keywords}[1]{\textbf{\textit{Keywords:}} #1}

\section{Introduction}
\begin{figure}[t!]	
	\centering
	\includegraphics[width=0.6\textwidth]{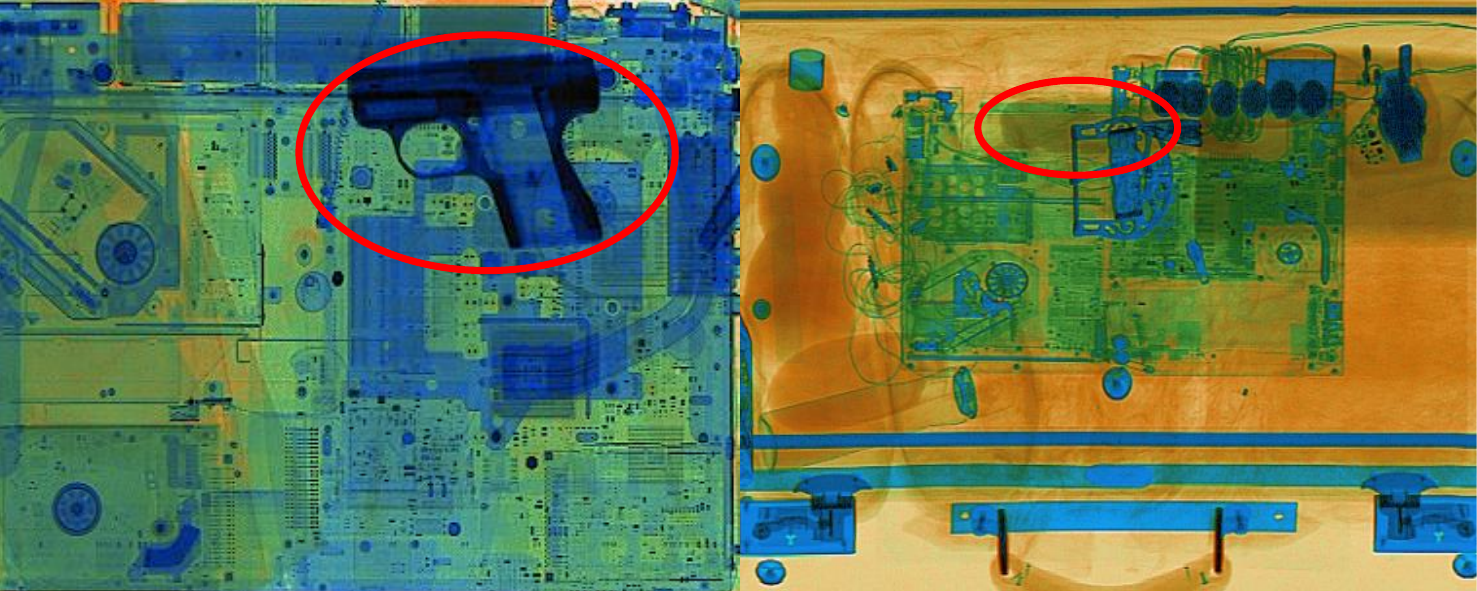}
	\caption[Threat objects in 2D X-ray images \cite{akcay2018using}.]{Threat objects in 2D X-ray images \cite{akcay2018using}. Left: threat object signatures of specific appearance (e.g., firearms) are easy to identify. Right: threat materials are more challenging to recognize in cluttered X-ray images.}
	\label{fig:x-ray}
\end{figure}

X-ray computed tomography (CT) scanners have been deployed at airports across the world to detect threat materials hidden in baggage. By inspecting the content in the 3D reconstructed image from the scanner, operators are able to identify potential threat objects without the need for manual search \cite{mouton2015review, hattenschwiler2018detecting}. Some threat objects such as firearms and knives can be recognized by their appearance, whilst threat (explosive) materials which can be made into threat devices could have varying physical appearances (e.g., different shapes and volumes, as illustrated for related work on 2D X-ray in Figure \ref{fig:x-ray}) \cite{mouton20143d, flitton2015object, hattenschwiler2018detecting}. Current approaches for material based threat detection, denoted as automatic threat recognition (ATR) or explosive detection systems (EDS) within the aviation security industry, may use dual energy computed tomography (DECT) \cite{mccollough2015dual} to perform this detection task based on a range of standardised threat material signatures. In addition, operators may perform a secondary manual inspection of the reconstructed 3D CT image to resolve materials highlighted by the ATR/EDS process.

Automatic threat recognition (ATR) intends to enable baggage screening more efficient and effective with only limited human intervention. Attempts have been made in recent works to address the ATR problem in 2D X-ray images \cite{bacstan2015multi, akccay2016transfer, mery2017modern, caldwell2017transferring, akcay2018using} and 3D CT images \cite{eger2011learning, wiley2012automatic, segmentationAlert, megherbi2013investigating, mouton20143d, mouton2015review, mouton2015materials, atrAlert, martin2015learning}. Most existing works, however, focus on the recognition of threat objects having specific shape-based appearances (e.g., firearms, bottles, knives, etc.), which can not be directly applied to the recognition of threat materials.
In this paper, we aim to detect threat materials in baggage based on 3D CT images. It is usually formulated to be a visual recognition problem \cite{atrAlert} in the X-ray CT imagery domain.  A typical approach to addressing this problem is to use machine learning techniques \cite{mouton2015review, atrAlert}. Specifically, the potential object volumes are firstly extracted from the CT images using a 3D segmentation algorithm, then the objects are classified as threats or non-threats using a binary classifier \cite{atrAlert}.

One limitation of conventional ATR is the lack of adaptability. Once the ATR models are deployed to the scanners, they work for the detection of threat materials pre-defined during the model training. In real scenarios, however, the definition of threat materials evolves with the changing nature of the international aviation security landscape and the threat actors involved. As a result, material signatures not originally defined as a threat can emerge as new threat materials \cite{liscouski2016evolving}.  Although such materials may be detected with current ATR, specific adaptations are required to improve both sensitivity and specificity of screening. Adapting the deployed ATR algorithms to these new material signatures is a non-trivial task. First, scanner manufacturers need to adjust their algorithms to tackle the new definitions which might involve new training data collection, annotation and algorithm re-training. In addition, it may take considerable time to have the new algorithm variants certified for deployment \cite{atrAlert}.

To address this issue, adaptive automatic threat recognition (AATR) is proposed \cite{to7,manerikar2018adaptive,paglieroni2018consensus} by the US Department of Homeland Security and the ALERT (Awareness and Localization of Explosives-Related Threats) Center of Excellence \cite{bhattacharjee2008new}.
The problem AATR aims to address differs from conventional binary classification problems in that the definition of threats and non-threats could vary from situation to situation, which raises the requirement for the algorithm to be adaptive to varying definitions of threat and non-threat material signatures. 
Conventional automatic threat recognition aims to classify an object within baggage as a threat or a non-threat deterministically without the capability of adapting to the change of threat definitions. However, an adaptive automatic threat recognition algorithm could take the definition of threats as an additional argument and classify an object as a threat or not accordingly. As a result, the output of AATR is determined not only by features of the object but also by the definition of threats.
To formulate the AATR problem, ALERT defines a series of adaptability metric (AM). Each adaptability metric consists of multiple objects requirement specifications (ORS). For example, the AM of varying materials aims to evaluate how well an AATR algorithm can adapt to varying materials. Further conceptual details of the AM and ORS defined by ALERT are described in \cite{to7} and related to our work in Section \ref{sect:evalAdapt}. 

In this paper, we present an approach to resolving the AATR problem originally posed by \cite{to7}. In the initial stage, we propose a multi-scale 3D image segmentation algorithm to segment objects of different scales and shapes in the baggage. Subsequently, we utilize a multi-class support vector machine (SVM) \cite{hsu2002comparison} to classify the segmented object signatures into different classes of material. Once the materials of segmented objects are known, our AATR approach can recognize potential threat objects according to the  material signatures specified by the requirements. Instead of assigning a class label to a given segmented object, we calculate its probability of belonging to each class. By adaptively manipulating these probabilities, we are able to make the classification results bias to some specific classes intentionally to satisfy varying probability of detection requirements. 


Contributions of this paper are summarized as follows:
\begin{easylist}[itemize]
	& A novel 3D segmentation algorithm is developed to separate multi-scale material objects within the CT baggage image so that each segmented object can be further classified as a threat or non-threat material signature;
	& An approach to the AATR problem, as originally posed by \cite{to7}, is proposed with the adaptabilities to varying specifications of threat material signatures, varying \textit{probability of detection} (PD) requirements and new unseen threat materials for which there are no examples available during algorithm training; 
	& An evaluation is conducted to validate the feasibility and effectiveness of the proposed solution to the AATR problem.
\end{easylist}
\section{Related Work}
\label{sect:related}
In recent years, 2D X-ray image based threat object recognition has been extensively studied \cite{akccay2016transfer,akcay2018using,miao2019sixray}, although even the use of multiple view X-ray suffers from the challenges of object recognition under varying orientation and inter-object occlusion. This can be addressed by 3D X-ray CT imaging which provides abundant information as a 3D volume comprised of successive, parallel X-ray image slices \cite{hattenschwiler2018detecting}. 3D X-ray CT imaging can be categorized into single-, dual- and multi-energy CT based on how many separate energy levels are used for scanning. For the purpose of material differentiation, it has been shown that dual- and multi-energy CT has an advantage over single-energy \cite{eger2011learning,mouton2015materials}. Since the main focus of this paper lies in the adaptability of an automatic threat recognition algorithm, we particularly focus on 3D object segmentation and material classification for automatic threat recognition in 3D baggage CT images. For a broader view on the background knowledge and techniques of automatic CT image processing for baggage screening, we refer readers to \cite{singh2003explosives,wells2012review} and \cite{mouton2015review}.

\subsection{CT Image Segmentation}
Although CT image segmentation has been extensively studied in literature \cite{megherbi2013investigating,lee2015review,mouton2015review}, most approaches are specially designed for medical CT images of organisms and thus not directly applicable to our task in which the threat signatures could have varying shapes and materials \cite{mouton2015review}. To promote the automatic threat recognition performance in baggage screening, the first initiative regarding 3D CT image segmentation was proposed via \cite{atrAlert, segmentationAlert}. The initiative resulted in effective CT image segmentation approaches which can be used for automatic threat recognition \cite{segmentationAlert}. Among the outcomes of this initiative, Grady et al. \cite{grady2012automatic} proposed a 3D CT image segmentation algorithm featured by a novel Automatic QUality Measure (AQUA) model that measures the segmentation quality for the segmented object so that good segmentation results can be achieved by learning segmentation parameters to optimize this measurement based objective. The model is learned based on 92 ``good quality" object segments by a data-driven machine learning approach. As a result, the effectiveness of the learned AQUA model might rely on the features (e.g., shape, geometric features) of the training data and is limited when there exist new objects of arbitrary shapes to segment. Wiley et al.\cite{wiley2012automatic} adapted a medical segmentation technique called Stratovan Tumbler \cite{wiley2012analysis} to the baggage CT image segmentation problem. Kernels of different parameter values are used to deal with the boundaries of objects having different shapes and sizes. Our segmentation algorithm employs a similar idea that multi-scale structuring elements (Section \ref{sect:shapesplit}) are used to segment objects of different sizes recursively.

Mouton et al. \cite{mouton2015materials} proposed a novel technique for the 3D segmentation of unknown objects in the baggage CT images. The proposed algorithm takes advantage of the appearance information of objects to be segmented (e.g., handguns and bottles) which, however, is not applicable to our problem since the concerned threat materials could appear in very different forms. Martin et al. \cite{martin2015learning} proposed a learning based framework for joint object segmentation and threat recognition from volumetric CT images. Their work was based on the dual-energy X-ray computed tomography which can provide more essential information to differentiate materials \cite{mccollough2015dual} while our work focuses on single-energy CT images.

\subsection{CT Image Based Material Classification}
One key element in automatic threat recognition is materials classification. CT image based material classification is used to distinguish the candidate signatures of threat materials from those of non-threat materials. 
Recently, a new initiative was proposed towards automatic threat materials recognition based on baggage CT images \cite{atrAlert}. Threat materials recognition in baggage CT images is usually formulated as a material classification problem based on the segmentation results. Ye et al. \cite{atrAlert} proposed a new 3D CT image segmentation approach followed by classifying the segmented candidate threat material signatures to be threats or non-threats using multiple SVM classifiers. These classifiers are trained on different clusters of training object signatures based on their shapes so that the threats and non-threats falling into different clusters can be recognized using different classifiers respectively. Evaluated on the particular dataset for this initiative, they achieved  95\% probability of detection with a probability of false alarms less than 10\%. Resulting from the same initiative, Zhang et al.\cite{atrAlert} used a pixel classification approach for CT image segmentation which can be implemented by an expectation-maximization (EM) algorithm and enhanced by using a Markov random field (MRF) to impose some spatial smoothness-constraints on the iterative two-step EM process. As for the classification of threats and non-threats, a collective classification algorithm was employed, where, one SVM classifier was trained for each type of material. Evaluated on the same data as Ye et al. \cite{atrAlert}, the performance is slightly worse yet still competitive with 89.2\% probability of detection and 9.7\% probability of false alarms. SVM classifier was also employed for CT image classification  by Flitton et al. \cite{flitton2015object} in which, however, the classification focuses on specific objects rather than materials.

The AATR problem to be addressed in this paper aims to improve the adaptability of conventional ATR algorithms. On one hand, we take advantage of the framework used in many works that consist of a 3D segmentation algorithm followed by classification. On the other hand, our proposed segmentation and classification methods make the least use of supervision information (especially the shapes of objects) to promote the generalization capability. More importantly, our approach has the adaptability to varying requirements in practical applications. In parallel to this work, Paglieroni et al. \cite{paglieroni2018consensus} proposed a spatial consensus relaxation method to determine the most likely material composition for each CT image voxel, followed by threat signature classification to address the same AATR problem.
\section{Method}
\label{sect:method}

\begin{figure}[t]	
	\centering
	\includegraphics[width=0.6\textwidth]{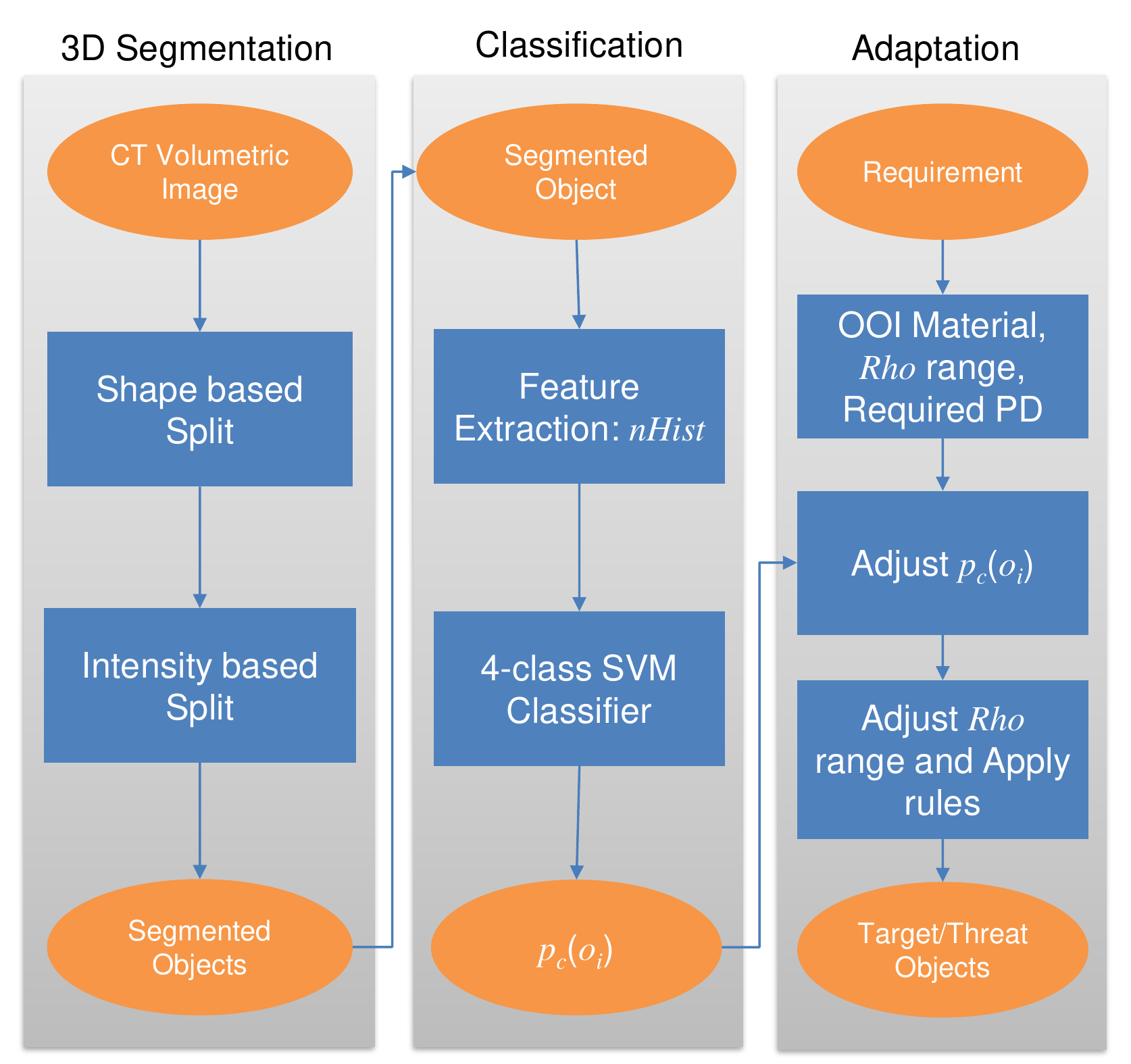}
	\caption{The framework of our proposed adaptive automatic threat recognition (AATR) approach. }
	\label{fig:framework}
\end{figure}

Our AATR approach consists mainly of three components as shown by the diagram in Figure \ref{fig:framework} (i.e. 3D segmentation, material classification and adaptation to varying requirements). These three components form the pipeline of our approach to threat recognition with the adaptability to varying requirements (e.g., objects of which properties are required to be recognized as threats). As the first step, 3D segmentation is used to separate all the object signatures in baggage based on the raw CT volumetric data. Given a 3D CT image as input, it is firstly binarised by voxel thresholding and we define these voxels as background. Within this context, the non-zero voxels are now defined as object volumes. The segmentation algorithm attempts to find out the 3D volume of each individual object signature which is characterized by a set of connected voxels of the same material intensity (i.e. voxel value within a CT image) range. Voxels of each segment are labelled with a unique integer label, while the background voxels are labelled with zeros. Once segmented objects have been obtained, we need to decide if the segment object is a threat or non-threat according to the definition of threat objects. Since the threat objects are defined in terms of their materials and physical properties such as the density (i.e. defined as the mean intensity of voxels of an object signature), a classification step is employed to recognize the material types of the segmented objects (i.e. \textit{saline}, \textit{rubber}, \textit{clay} and \textit{others}). As a result, any segmented object in the image can be classified as a threat or non-threat according to its material type derived from the classification result and density which can be calculated. Due to the inaccuracy of segmentation and classification results, the detection performance is not perfect with the \textit{probability of detection} (PD) less than 100\% along with a non-zero \textit{probability of false alarm} (PFA). To satisfy the particular requirements in varying practical scenarios, we employ adaptation techniques to trade off the probability of detection and false alarm. For instance, we take measures to have more segmented objects classified as threats, which will inevitably mis-classify some non-threats as threats (increasing the probability of false alarm ) though more true threats are expected to be classified correctly (increasing probability of detection) at the same time. As a result, the probability of detection of a specific material can be adaptively increased or decreased to satisfy the requirements.  In our approach, the adaptation can be implemented by manipulating the output probabilities of different classes (i.e. material types) from a support vector machine (SVM) classifier.

\subsection{3D Segmentation}
\label{sect:segmentation}
Image segmentation is an important pre-process in many computer vision tasks. By segmenting objects in images, we are able to focus on the visual information of the individual object without interference from the background and other objects. Image segmentation has been studied extensively in recent decades, achieving good performance by learning from a large amount of training data, using the heterogeneous features of the objects such as color, shape, texture and semantics. However, the techniques commonly used for image segmentation are not readily applicable to the 3D CT images in our problem. First, the CT images are 3D volumetric data while most state-of-the-art image segmentation algorithms are designed for 2D images. Second, the objects in the CT images are characterized only by the voxel intensities, lacking of the key information such as color and texture in the natural images. Finally, the objects concerned in our problem could have arbitrary shapes without the prior semantic knowledge.
Therefore we propose a novel 3D segmentation algorithm to segment objects of varying materials in 3D CT images. Our segmentation algorithm consists of two steps: shape based split and intensity based split. In the shape based split, the algorithm takes advantage of the morphological operations and the connected-component  labelling (CCL) \cite{rosenfeld1966sequential} method to split the objects which are not or slightly touching one another. Subsequently, the intensity based split is employed to handle the merged objects on which the first step has failed but the merged objects are of different materials.
\subsubsection{Shape Based Split}
\label{sect:shapesplit}
The basic idea of the shape based split is to use the classical connected-component labelling (CCL) method to label objects which are not connected. The objects not connected to any other objects can be directly segmented from others. Objects in baggage, however, are usually compactly packed, touching each other thus posing a big challenge to the original CCL method. To address this issue, we apply morphological opening operations to split touching objects. We first apply erosion to remove the connection voxels among multiple objects, then the CCL method can be applied to label the individual objects. Finally, the dilation operation is applied to the segmented individual objects to compensate the voxel loss caused by erosion operations by using the same structuring element as used for erosion.

\begin{figure*}[h]	
	\centering
	\includegraphics[width=\textwidth]{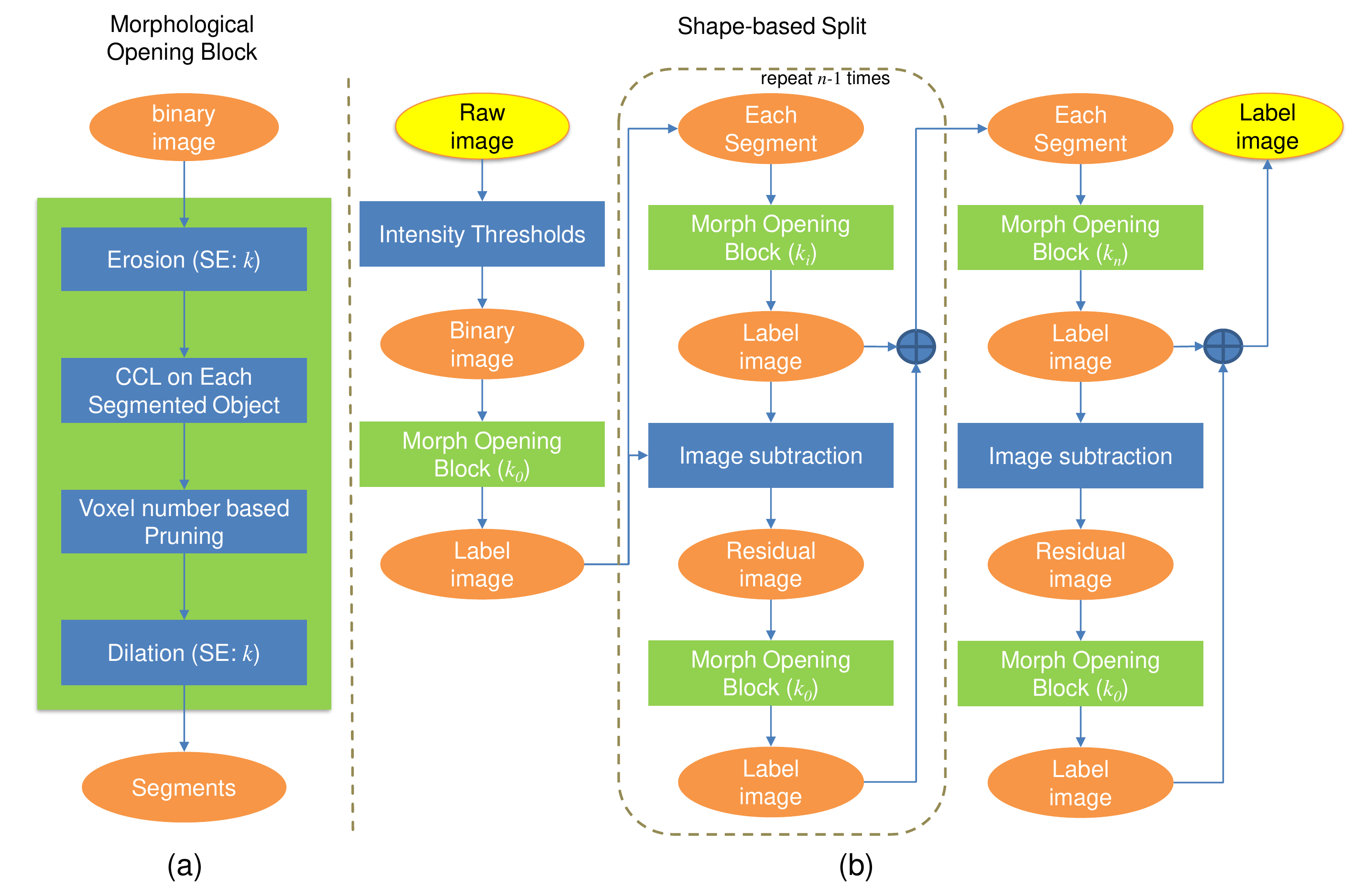}
	\caption[The proposed shape-based split using multi-scale morphological opening operations.]{The proposed shape-based split using multi-scale morphological opening operations. (a) The morphological opening operation with the structuring element parameter $k$. (b) Multi-scale morphological opening work-flow.}
	\label{fig:shapesplit}
\end{figure*}

One key factor in the morphological opening operation is the selection of the structuring element (or the kernel). The parameterisation of the structuring element (e.g., the radius of a sphere) used for erosion determine what kind of connections between two objects can be broken down. A small-scale structuring element can only split objects connected by a small number of voxels, while a large-scale structuring element has the risk of completely removing small objects. To solve this problem, we use a novel multi-scale morphological opening approach, using multiple structuring elements of different parameter values recursively. Our multi-scale morphological opening approach is shown in Figure \ref{fig:shapesplit}. It can split objects connected by a large number of voxels without mistakenly removing small objects.

Figure \ref{fig:shapesplit}(a) shows the basic morphological opening operation used in the multi-scale morphological opening approach. The workflow of the multi-scale morphological opening is illustrated by the diagram in Figure \ref{fig:shapesplit}(b). To facilitate our discussion, we describe some notations and terms used in the following content. A \textit{raw image} is defined as a three-dimensional CT image having continuous voxel values. A \textit{binary image} is defined as a three-dimensional volumetric image with binary voxel values (i.e. $0$ indicates the background voxels and $1$ indicates the foreground voxels). A \textit{label image} $I^L$ is defined as a three-dimensional volumetric image with voxel values $v\in \{0,1,...,n\}$, where $n$ is the number of labelled/segmented objects, $0$ indicates the background voxels and the values of $i$ indicates the voxels belonging to the $i$-th segmented object. Therefore a binary image is actually a label image with only one labelled object in this sense. 

The morphological opening block takes a label image $I^L, v\in \{0, 1, 2,..., n\}$ as input and output another label image $I^{L'}, v'\in\{0, 1, 2,..., n, n+1,...,n+m\}$, where $m\ge0$. The morphological opening operation is applied to each object specified in the input label image. Different from the classical morphological opening with only erosion and dilation operations, we add the voxel number based pruning process to remove the segmented objects having too few voxels and use the CCL method to identify segmented objects in the image. For any object $o_i = \{v| v=i\}$ in the original label image, if it can be further split into more than one objects after the morphological opening, the voxels of these segmented objects will be relabelled using different values respectively, otherwise the object will keep unchanged in the output label image. Particularly we keep the object unchanged if there is no segmented object left after the morphological opening operation.

Now we describe the multi-scale morphological opening approach for the shape-based split in detail. Given a raw image $I$ and its voxel value $v\in \mathbb{R}$, our goal is to obtain a label image $I^L, v \in \{0, 1, 2, ..., n\}$, specifying $n$ segmented objects, and the segmented object $o_i$ is specified by a set of connected voxels whose values are equal to $i$. First, the raw image $I$ is transformed to a binary image by voxel-wisely comparing its voxel values with a lower threshold $t_{lower}$ and an upper threshold $t_{upper}$ which are determined by the prior knowledge, so that the objects of unconcerned materials can be directly removed. The binary image, which is a special case of the label image defined above, is then fed into the morphological opening block where the parameter of the structuring element used for erosion is set as $k_1$. The resultant label image $I^L_1$ contains $n_1$ segmented objects. Second, we apply the morphological opening block again on the label image $I^L_1, v\in\{0, 1, 2, ..., n_1\}$ with a greater structuring element parameter $k_2>k_1$. As a result, a label image $I^L_2$ containing $n_2\ge n_1$ segmented objects can be obtained. However, there exists a risk of mistakenly removing small/thin objects by applying the morphological opening with a relatively large $k_2$. For example, assuming there is an object $o_i$ labelled in the label image $I^L_1$ which is merged by two objects, and one of them happens to be small or thin such that it is completely removed after being split. To avoid this erroneous occurrence, we reinspect the voxels removed from the label image $I^L_1$ to recover the mistakenly removed objects if there are any. Specifically, we apply an \textit{image subtraction} operation to get the \textit{residual image} $I^R$, a binary image characterizing the background difference between label images $I^L_1$ and $I^L_2$,
\begin{equation}
\label{eq:resImg}
I^R = I^L_1 - I^L_2, v^R \in \{0, 1\}
\end{equation}
where any voxel value $v^R$ in the residual image $I^R$ is computed by:
\begin{equation}
\label{eq:resVox}
v^R = \begin{cases}
0 &\text{if } v^L_1>0, v^L_2>0 \text{ or } v^L_1=v^L_2=0,\\
1 &\text{if } v^L_1>0, v^L_2=0.
\end{cases}
\end{equation} 
The residual image $I^R$ is re-inspected by applying the morphological opening operation with the same structuring element parameter $k_1$ as that used in the first step. The value of $k_1$ is selected so that none of the threat material signatures can be mistakenly removed by the erosion operation with such a small structural element. The resultant label image is denoted as $I^{LR}_2, v\in \{0, 1, ..., n_R\}, n_R\ge 0$. We update the label image $I^L_2$ by adding the residual label image $I^{LR}_2$:
\begin{equation}
\label{eq:addImg}
I^{L'}_2 = I^L_2 + I^{LR}_2, v^{L'}_2 \in \{0, 1, 2, ..., n_2+n_R\},
\end{equation}
where any voxel value $v^{L'}_2$ in the label image $I^{L'}_2$ is computed by:
\begin{equation}
\label{eq:addVox}
v^{L'}_2 = \begin{cases}
v^L_2 &\text{if } v^{LR}=0,\\
n_2 + v^{LR}_2 & \text{otherwise}.
\end{cases}
\end{equation}
Once the label image $I^L_2$ is updated, it can be fed into another morphological opening block with a greater structuring element parameter $k_3>k_2$ and repeat the steps again to further split the objects specified in the label image $I^{L'}_2$ as shown in Figure \ref{fig:shapesplit}(b). 

Technically, the multi-scale morphological opening operations can be repeated many times with gradually increased structuring element parameters $k$.
The depth of such recursion is subjective to the scales of objects-of-interest in the CT volumes and the constraint of computation time. On one hand, more rounds of recursion are needed if the scales of the objects-of-interest vary significantly for good segmentation performance. On the other hand, more rounds of recursion will require more computation. As a result, the depth of recursion can be decided by trading-off required segmentation performance against computation. In our experiments, we recursively apply the morphological operations for three rounds and empirically set $k_1=2$, $k_2=3$ and $k_3=8$.
\subsubsection{Intensity Based Split}
\label{sect:intensitysplit}
\begin{figure}[h]	
	\centering
	\includegraphics[width=0.8\textwidth]{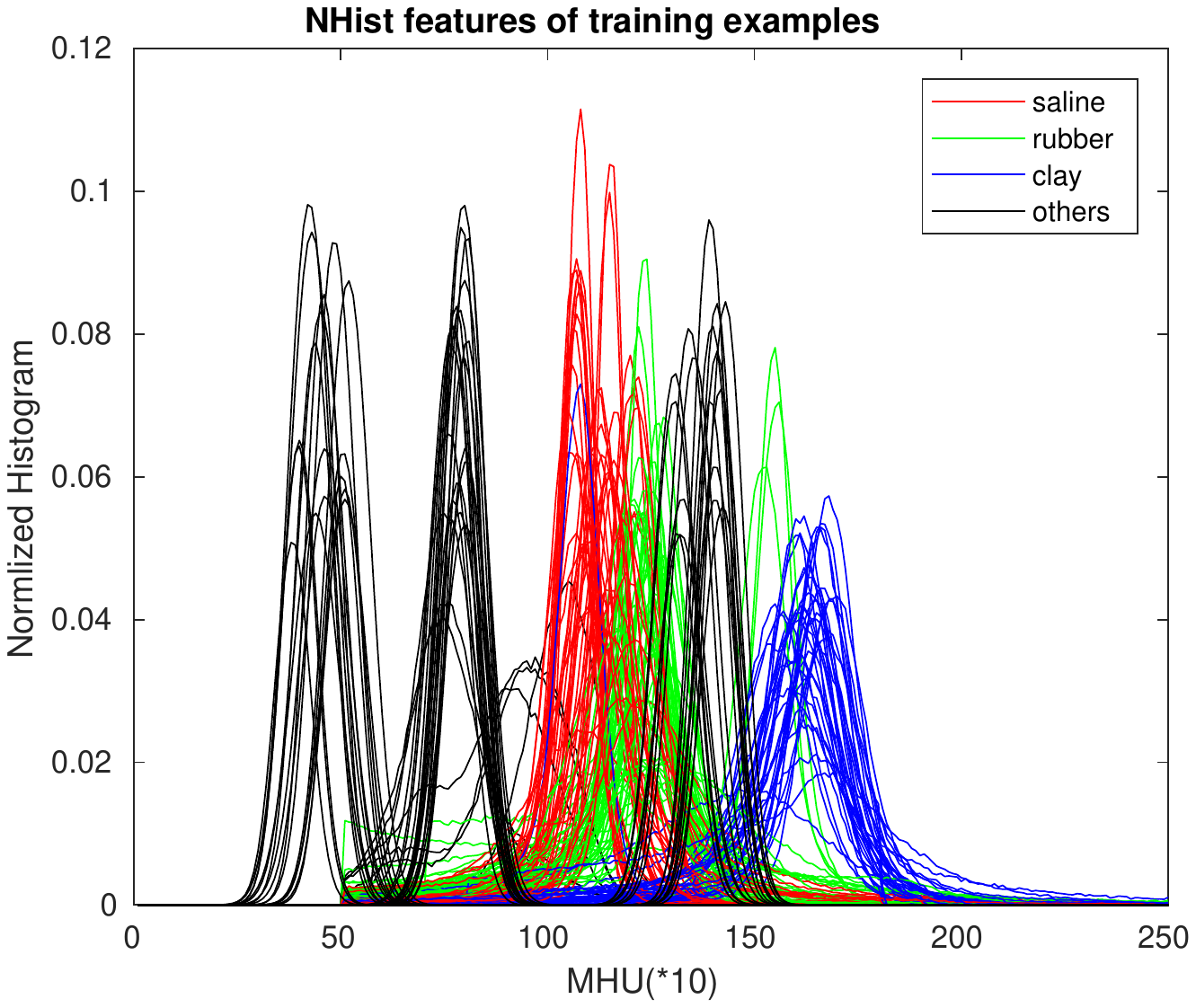}
	\caption{The normalised histogram features of different materials.}
	\label{fig:intensityDist}
\end{figure}

The shape based split method can fail when the merged objects are compactly touching each other. The intensity based split method provides a complementary way to address this issue when the merged objects happen to have distinguishable intensity distributions. Objects of different materials have different intensity distributions as shown in Figure \ref{fig:intensityDist}. When the merged objects are of different materials, the voxel intensity will be distributed in multiple ranges corresponding to the intensity range of each material. Based on this idea, the intensity based split aims to discover the potential materials in a merged object and assign respective labels to voxels of the merged object.

\begin{figure}[h]	
	\centering
	\includegraphics[width=0.7\textwidth]{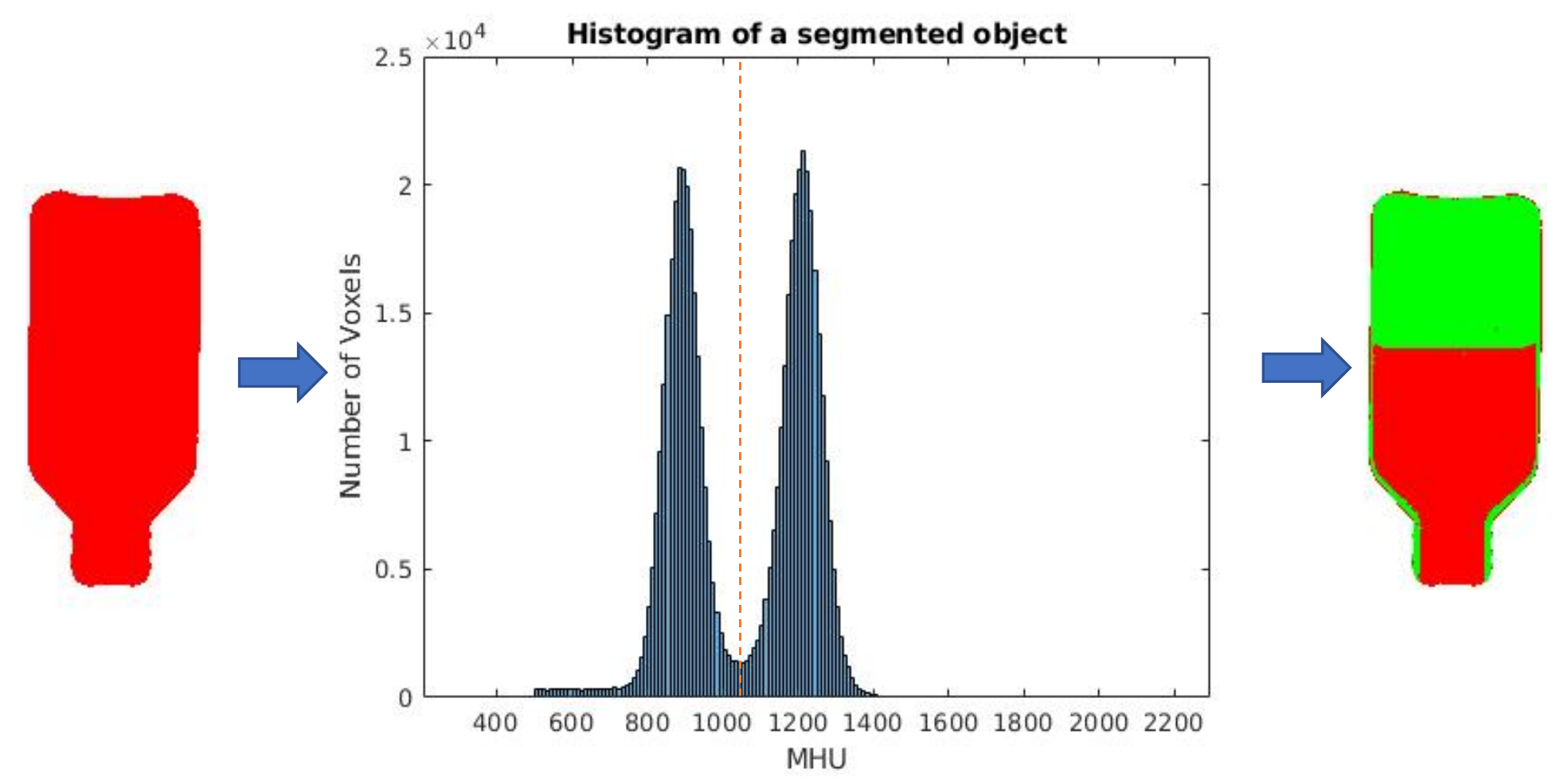}
	\caption{An illustration of intensity based split.}
	\label{fig:intensitysplit}
\end{figure}

A variety of enabling techniques could be used for our purpose of intensity based split. For example, a probabilistic clustering algorithm using Expectation-Maximization (EM) can be employed to model the intensity distributions of different materials which are then used to label each voxel of a given volume as one of the materials  \cite{simanovsky2009method}. Due to the variations of intensity distributions for the same material (c.f., Figure \ref{fig:intensityDist}), the estimated models using EM might not generalize well to some specific new volumes. Instead of modelling the intensity distributions for all materials, we employ a peak-detection method that explores the volume-specific intensity distribution information and does the segmentation.

Specifically, given a segmented object $o_i$ in the label image $I^L$ generated by the shape based split, we extract the intensity values of all the voxels belonging to this object from the raw image $I$. The histogram of the intensity can be computed and gives an intensity distribution curve as shown in Figure \ref{fig:intensitysplit}. Subsequently, we use a peak detection algorithm to search the number of objects residing in the segmented object $o_i$ and the intensity range for each of them. The peak detection algorithm uses a sliding window to locate a peak whenever the value in the middle of the sliding window is the maximum. Once the peaks are detected, we search the lowest value between two peaks as the intensity boundary of the two objects. If only one peak is detected, it means there is only one type of material in this segmented object, and we leave it unchanged. Otherwise, we assign the voxels whose intensity values are in the same intensity range to a new object label. As a result, the merged object $o_i$ can be split into multiple objects of different materials. The whole procedure of the intensity based split is illustrated in Figure \ref{fig:intensitysplit}.

\subsection{Material Classification}
\label{sect:classification}
The 3D segmentation algorithm creates a label image $I^L$ for each baggage specifying a collection of object signatures, among which there are both material signatures of interest (threats) and others (non-threats). The existence of non-threats in the label image will incur false alarms. To get rid of the false alarms, we need to distinguish the threat signatures from the non-threat items. In this section, we first describe how the classification is employed in the traditional ATR and then extend it to the AATR problem.

In the conventional automatic threat recognition problem, the threat materials are pre-defined and fixed, we only need to annotate the signatures as threats or non-threats by a binary classifier. The binary classifier can be trained using training data in which all the segmented objects are annotated as threats or non-threats according to the definition. For example, the materials of \textit{saline}, \textit{rubber} and \textit{clay} are defined as threat materials in our experiments, we annotate all the signatures of these three materials in the training data as threats and other objects as non-threats. To train a classifier for this purpose, we need to extract useful features from the segmented objects. Based on the definition of threats (i.e. materials of \textit{saline}, \textit{rubber} and \textit{clay}), we use the normalised histogram of intensities as the feature for classification since it is able to characterize the material type of an object. 

In the adaptive automatic threat recognition problem, it is required to recognize the material type of objects so that the detection decision can be made adaptively to varying requirements of threat material recognition. To this end, we train a four-class classifier, \{``\textit{saline}", ``\textit{rubber}", ``\textit{clay}", ``\textit{others}"\}, where the class ``\textit{others}" denotes a collection of unknown materials other than the three known materials. Again, we use the normalised histogram of intensities as features for classification. We use support vector machine (SVM) with a linear kernel as the classifier in our experiments, though other types of classifiers such as random forest \cite{breiman2001random} and neural networks are also applicable.

The SVM classifier is trained using features of ground truth object signatures in the training data for known material classes (i.e. \textit{saline}, \textit{rubber} and \textit{clay}). For the ``\textit{others}" class, we synthesize a collection of features as training data belonging to this class. Specifically, a Gaussian function $g(x; a,\mu,\sigma)$ is used to synthesize a feature vector $\bm{v}$:
\begin{equation}
\label{eq:gaussian}
\bm{v} = g(x;a,\mu,\sigma)= a\cdot e^{\frac{-(x-\mu).^2}{\sigma^2}},
\end{equation}
where $\mu$ is randomly selected only if it is out of the density ranges of three known materials, $\sigma$ is a random number chosen from an empirical range derived from the intensity distribution of the ground truth material signatures in the training data (see Figure \ref{fig:intensityDist}).

\subsection{Adaptation to Varying Requirements}
\label{sect:adaptation}
The adaptation step aims to enable our approach to adapt itself to varying requirements in practical scenarios. Specifically, we focus on the adaptability metrics defined by \cite{to7} in terms of: (a) varying threat materials; (b) varying probability of detection (PD) requirements.

Given a segmented object, the classification result allows us to know which class it belongs to. With this information, it is straightforward to classify the segmented objects whose material types are not included in the requirement specification as non-threats. To enable the adaptability, we also need to know the probability of belonging to each class for a given segmented object. The probability estimation method for SVM proposed in \cite{wu2004probability} is used in our experiments. 
We trade off the probability of detection and the probability of false alarm by manipulating the classification results thus adapting our approach to satisfying the requirements of PD/PFA. Specifically, we adjust the probability of classes $p_c(o_i)$ for a given segmented object $o_i$ as follows:
\begin{equation}
\label{eq:adjustProb}
{p'}_c(o_i) = p_c(o_i) + f_c(t_{PD}), c\in\{saline, rubber, clay, others\}
\end{equation}
where $f_c(t_{PD})$ is an \textit{offset} used to adjust the output probability of the classifier for class $c$, and $t_{PD}$ is the required probability of detection for a specific material class. Intuitively, when the offset for class $c$ is positive, the objects will have higher probabilities to be classified as class $c$. On the other hand, if the offset is negative, chances are less to classify the objects as this class.

The key issue is how to get the offset functions $f_c(t_{PD})$ for different classes. We repeat experiments using training data by setting the offset to be varying values and calculate the corresponding probability of detection. With pairs of probability of detection and offset values, we can fit the data using smoothing splines \cite{de2005spline} to obtain the offset functions $f_c(t_{PD})$. Subsequently, these functions can be used to compute the offset values given the required probability of detection $t_{PD}$ for a specific class.

Other than the manipulation of output probabilities by Eq. (\ref{eq:adjustProb}),  we can also trade off the probability of detection and the probability of false alarm by adjusting the specified density range. In real scenarios, threat material signatures are defined with additional information of the density range, i.e., a threat signature of a specific material should have a density within the pre-defined density range in the requirement specification. The density, $Rho$, is the mean intensity over all voxels of an object. Due to the inaccuracy of segmentation and noises in the raw CT data, the computed densities of segmented objects could have biases. With this regard, we adjust the density range by a factor $\alpha$ in the following way:
\begin{equation}
\label{eq:alpha}
Rho'_{min} = Rho_{min}*\alpha,\\
Rho'_{max} = Rho_{max}/\alpha,
\end{equation}
where a larger $\alpha$ leads to a smaller density range and thus a lower probability of detection and a lower probability of false alarm. Since the parameter $\alpha$ in Eq. (\ref{eq:alpha}) plays a similar role as the \textit{offset} in Eq. (\ref{eq:adjustProb}), we will find and fix the optimal values of $\alpha$ and adjust the \textit{offset} values in our experiments to enable the adaptability to varying probability of detection requirements.
\section{Experiments and Results}
\label{sect:experiment}
To validate the effectiveness of our proposed approach, we conduct a series of experiments on real baggage CT image data. In this section, we first describe the CT image data and evaluation metrics used in our experiments. Subsequently, we describe the experiments designed to evaluate different components (i.e. segmentation, classification and adaptation) of the proposed approach. Experimental results are presented in each subsection following the description of experiment settings.
\subsection{Experimental Data}
\label{sect:data}
We use data provided by ALERT \cite{to7} in all the experiments. The baggage are scanned with a medical CT scanner Imatron C-300. The image size is $512\times512$ corresponding to the $475\ mm \times 475\ mm$ field of view. The slice spacing is $1.5\ mm$, leading the voxel volume to be $(475/512)^2*1.5=1.291\ mm^3$. The range of the voxel values is $0\rightarrow32,767$ MHU, where the value of air is 0, and the value of water is 1024. A public dataset (ALERT TO4 \cite{atrAlert}) and a sequestered dataset (ALERT TO7 \cite{to7}, unseen) collected by ALERT are employed in our experiments. The public dataset consists of 188 baggage CT images in which there are 446 object signatures of three materials (i.e. \textit{saline}, \textit{rubber} and \textit{clay}). These object signatures could be threats or non-threats according to the definitions of threat in different situations. For example, in one of our experiments, \textit{saline} is the only material defined as threat, as a result, all the \textit{rubber} and \textit{clay} signatures will be non-threats. Apart from the material, other physical properties of the objects such as \textit{mass} and \textit{density} are also used to define the threat. The definition of threat will be clarified in each experiment. The ground truth voxels are labelled by ALERT for all the objects which could potentially be threats. These ground truth annotations of object signatures in CT images can be used for the evaluation of our segmentation and classification algorithms described in Section \ref{sect:method}. The sequestered dataset is used for testing the adaptability of AATR approaches. The CT images in the sequestered dataset consist of not only three known materials in the public dataset, but also some new materials not seen in the training stage. Since the sequestered dataset is not openly available to the authors, most of the experiments presented in this paper are conducted based on the public dataset, though some experimental results on the sequestered data are also provided with testing on this unseen dataset by the ALERT team, independent from the authors. We divide the 188 baggage CT images in the public dataset into two subsets (i.e. the odd set and the even set) by the baggage scanning serial number. This leads to 94 images in the odd set and 94 images in the even set. We use one subset for training and test our approach on the other subset. Our use of both a public and sequestered dataset in this manner directly mimics the unknown nature of the AATR task under deployment conditions against an unknown threat adversary.
\subsection{Evaluation Metrics}
\label{sect:metric}
We employ the evaluation metrics defined by ALERT \cite{to7} in our experiments. Segmentation performance is evaluated by measuring how many ground truth object signatures in the test data can be matched with the segmented objects. Here the concept of \textit{match} is defined by the precision $P$ and the recall $R$. Suppose we use $G$ and $S$ to denote the voxel sets of a ground truth signature and a segmented object respectively, the precision and recall can be obtained by:
\begin{equation}
\label{eq:pr}
P = \frac{volume(G\cap S)}{volume(S)},
R = \frac{volume(G\cap S)}{volume(G)}.
\end{equation}
A ground truth object signature $G$ is thought of as correctly segmented if any segmented object $S$ achieves $P\ge 0.5$ and $R\ge 0.5$ for bulks or $P\ge 0.2$ and $R\ge 0.2$ for sheets.

Once the matching results are obtained for all segmented objects, probability of detection (PD) and the probability of false alarms (PFA) are employed to evaluate the performance. Probability of detection is the ratio of the number of detected signatures to the number of ground truth signatures in all the CT images, whilst probability of false alarm is the ratio of the number of falsely detected non-threat signatures to the total number of non-threats in the CT images:
\begin{equation}
\label{eq:pdpfa}
PD = \frac{N_{detections}}{N_{threats}},
PFA = \frac{N_{false\ alarms}}{N_{non-threats}}
\end{equation}

\subsection{Evaluation of 3D CT Segmentation}
\label{sect:evalSeg}
To thoroughly evaluate the performance of our proposed segmentation algorithm, we conduct an ablation study to investigate the contribution of each component (e.g., shape based split, intensity based split) in the segmentation algorithm. In addition, we also investigate how different structuring element parameter values affect the segmentation performance. Specifically, we conduct a comparative experiment by comparing different baseline methods with our full model. The baseline methods used in our experiment are as follows:
\begin{easylist}[itemize]
	& \textbf{Baseline 1} uses only the morphological opening ($k=2$) with CCL.
	& \textbf{Baseline 2} uses only the morphological opening ($k=3$) with CCL.
	& \textbf{Baseline 3} uses only the morphological opening ($k=8$) with CCL.
	& \textbf{Baseline 4} uses only the shape based split.
	& \textbf{Our approach} uses shape based split followed by intensity based split.
\end{easylist}
The segmentation performance is evaluated by matching the segmented object signatures with the ground truth. In this experiment, we define all of 446 \textit{saline}, \textit{rubber} and \textit{clay} objects as threats regardless of their physical properties. A good segmentation algorithm is expected to successfully segment all these objects from the 3D CT images, which will lead to a high probability of detection. Inevitably, the segmentation algorithm will also segment objects other than the 446 objects of interest, which will contribute to the probability of false alarm. However, in this stage, we focus more on the probability of detection as an evaluation of segmentation performance and leave the false alarm issue to the next step of classification. The results of our approach and different baseline methods are shown in Table \ref{table:seg}. 

\begin{table}[!t]
	\centering
	{
		\centering
		\caption[]{Segmentation performance of our algorithm and baselines.\\
		}
		\label{table:seg}
		\begin{lrbox}{\tablebox}
			\begin{tabular}{c|ccc|cc|c|c}
				\hline
				\multirow{2}{*}{\textbf{Method}} &  \multicolumn{6}{c|}{\textbf{PD}} & \textbf{PFA}\\ \cline{2-8}
				&\textbf{saline} & \textbf{rubber}   & \textbf{clay}  & \textbf{bulk} & \textbf{sheet} &\textbf{overall} & \textbf{overall} \\
				
				\hline
				baseline 1 (k=2) & 0.55 & 0.69 & 0.55  & 0.55 & 0.71& 0.60 & 0.31 \\
				baseline 2 (k=3) & 0.60 & 0.64 & 0.51 & 0.59 & 0.58 & 0.62 & 0.30 \\
				baseline 3 (k=8) & 0.58 & 0.13 & 0.12  & 0.38 & 0.10 & 0.29& \textbf{0.21} \\
				baseline 4 (shape-split)& 0.80 & 0.92 & 0.73  & 0.78 & 0.91& 0.83 &0.49 \\
				ours			 & \textbf{0.90} & \textbf{0.96} & \textbf{0.97}  & \textbf{0.93} & \textbf{0.95} & \textbf{0.94}& 0.56 \\
				\hline
			\end{tabular}
		\end{lrbox}
		\scalebox{0.8}{\usebox{\tablebox}}
	}
\end{table}

The segmentation performance shown in Table \ref{table:seg} clearly indicates the effectiveness of our proposed 3D segmentation algorithm. The best overall probability of detection (0.94) is achieved by our full method.  Comparing the performance of our full method with that of different baseline methods, we know that each component in our method plays an important role in the segmentation. Baseline methods with only one uniform structuring element parameter value perform poorly as shown by baseline 1-3 with overall probability of detection as 0.60, 0.62 and 0.29 respectively. Baseline 1 and 2 using a small structuring element ($k=2, 3$) fail to split objects touching each other, while the baseline 3 with a big structuring element ($k=8$) mistakenly removes small threat objects. Finally, a comparison between baseline 4 (0.83 probability of detection) using only shape-based split and our approach (0.94 probability of detection) using both shape and intensity based splits provides evidence that the intensity based split is complementary to the shape based split and able to further improve the segmentation performance. 

\begin{figure}[t!]	
	\centering
	\includegraphics[width=0.5\textwidth]{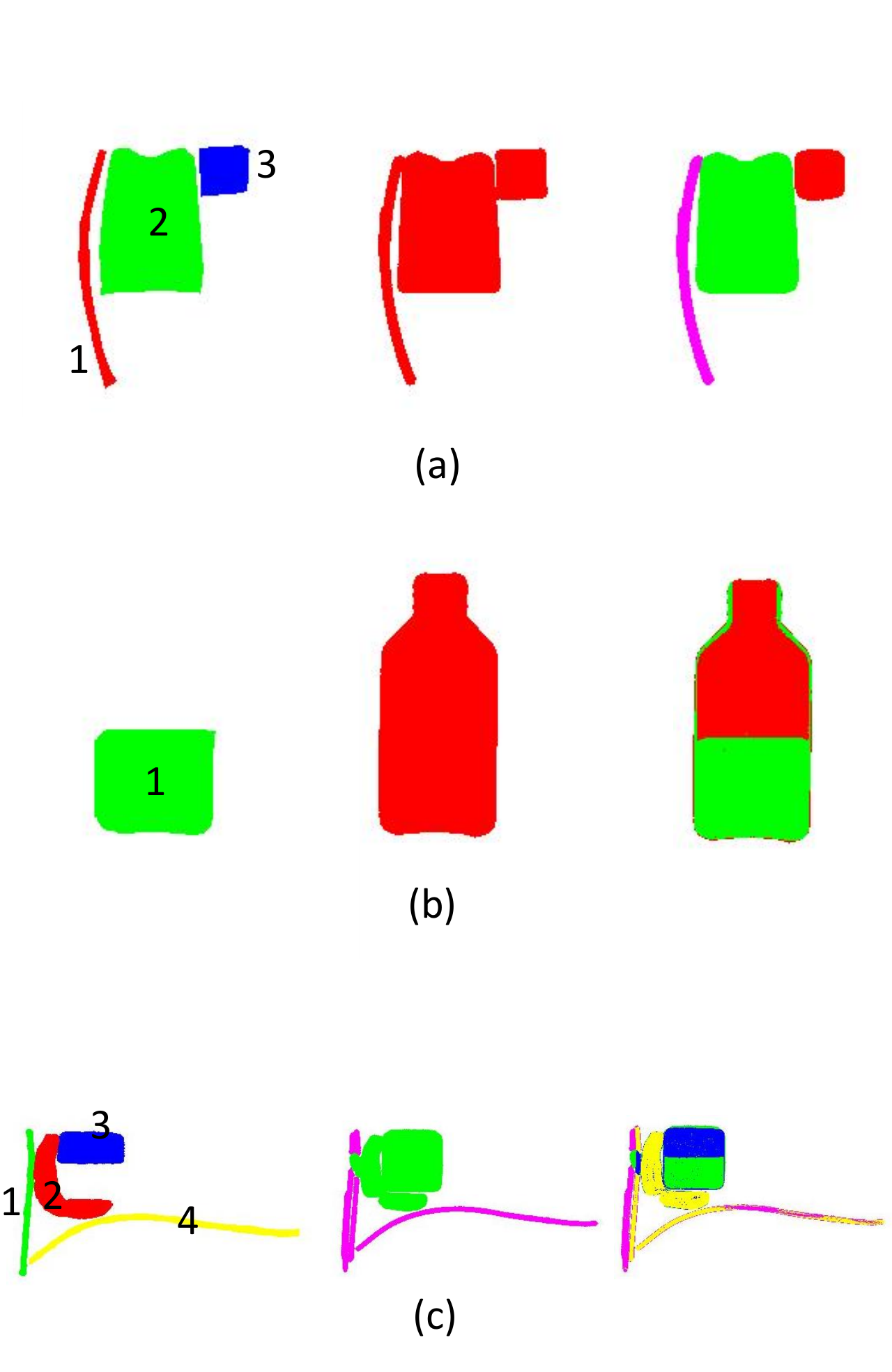}
	\caption[Exemplar segmentation results.]{Exemplar segmentation results. Each sub-figure represents the segmented object signatures in one slice of the 3D CT image. From left to right are the ground truth object signatures (only \textit{saline}, \textit{rubber} and \textit{clay} materials are annotated), segmented results with one of the baselines and segmented results with our approach respectively. Different colours are used to specify the separated object signatures. (a) The three merged objects cannot be split using a small structuring element $k=3$ (middle) but segmented well when using our multi-scale segmentation algorithm (right). (b) A half bottle of \textit{saline} can only be segmented by the intensity-based split (right). (c) The intensity-based split could cause inaccuracy when the merged objects have similar voxel intensities (objects 2 and 4 are not well separated even by our approach).}
	\label{fig:segExemplar}
\end{figure}

To give an intuitive insight of how the segmentation works for different cases, we present some exemplars in Figure \ref{fig:segExemplar}. The ground truth object signatures, segmented results with some baseline and segmented results with our approach are shown in the left, middle and right side of each sub-figure respectively. Each segmented object signature is indicated by one color for better visualization. Figure \ref{fig:segExemplar} (a) presents an example of three merged objects touching each other, which fails the shape based split when a small structuring element ($k=3$) is used, but can be successfully segmented when using a bigger structuring element ($k=8$). This lends us evidence that the multi-scale segmentation is effective and necessary. Figure \ref{fig:segExemplar} (b) and (c) illustrate two cases where the intensity based split is helpful. Figure \ref{fig:segExemplar} (b) presents an example of \textit{saline} in a bottle, for which shape base split successfully segments the bottle whilst the intensity based split can further identify the \textit{saline} signature. Figure \ref{fig:segExemplar} (c) shows the effectiveness as well as the limitation of our intensity based split when merged objects have similar intensities. One can see that the figure on the right side exemplifies the inaccurate segmentation of the object colourized as \textit{yellow} and \textit{pink}.

Overall, our proposed algorithm is able to segment objects of varying scales and shapes. Using only morphological operations and intensity-based split, our segmentation performance is comparable with other more complex approaches in \cite{atrAlert,manerikar2018adaptive,paglieroni2018consensus}. In addition, our segmentation method is not based on training hence less likely to suffer from overfitting issues than other learning based methods.
It is also noteworthy that a high probability of false alarm is expected when obtaining a high probability of detection. Therefore a further classification is necessary to reduce the false alarms resulted in the segmentation process.

\subsection{Evaluation of Material Classification}
\label{sect:evalClas}
We first investigate the performance of different classifiers in our material classification task. Considering the algorithm complexity and the capability of calculating conditional probability, we choose $k$ Nearest Neighbour ($k$NN), Neural Networks (NN) and kernel SVM in the comparative experiment. We use a radial basis function (RBF) in the kernel SVM method. The experiment is conducted using both true and synthetic object features, as a result, a multi-class classification problem with four classes (i.e. \textit{saline}, \textit{rubber}, \textit{clay} and \textit{others}) is formulated. We use cross validation and report the mean classification accuracy and standard deviation in Table \ref{table:classifier}. The Neural Network model performs the best, however, the class conditional probability output by this classifier cannot be adjusted to fulfill our purpose of Eq.(\ref{eq:adjustProb}) since the calculated class conditional probability is usually very close to either 0 or 1. As for $k$NN model, it performs the second best when $k=3$ which, however, makes it difficult to calculate the class conditional probability required by Eq.(\ref{eq:adjustProb}). As a trade-off, we use the linear SVM to classify materials in our approach.

\begin{table*}[!t]
	\centering
	{
		\caption[]{A comparison of different classifiers for material classification.\\
		}
		\label{table:classifier}
		\begin{lrbox}{\tablebox}
			\begin{tabular}{c|c|c|c|c|c}
				\hline
				\textbf{Method}&\textbf{$k$NN ($k$=3)} &  \textbf{$k$NN($k$=10)} & \textbf{SVM(linear)} & \textbf{SVM(RBF)} & \textbf{NN} \\ 
				\hline
				Accuracy & 92.9 $\pm$ 1.9  & 90.3 $\pm$ 1.6 & 91.2$\pm$ 1.4 & 86.7$\pm$ 2.2 & 93.5$\pm$1.1\\
				\hline
			\end{tabular}
		\end{lrbox}
		\scalebox{0.8}{\usebox{\tablebox}}
	}
\end{table*}

To evaluate the classification performance for conventional ATR and our adaptive ATR (AATR) respectively, we design two more experiments. The first experiment aims to evaluate conventional ATR, in which we assume the threat materials are pre-defined (i.e. all of the 446 \textit{saline}, \textit{rubber} and \textit{clay} objects are threats) and the training data for all types of materials are available. A binary SVM classifier is trained and used to classify the segmented objects to be either threats or non-threats. In the second experiment, we use a multi-class classifier to recognize different types of materials (i.e. \textit{saline}, \textit{rubber}, \textit{clay} and \textit{others}) where all the unknown materials are collectively treated as one class ``\textit{others}". To compare with the binary classification, the same definition of threats has been used in this experiment, i.e., all of the 446 \textit{saline}, \textit{rubber} and \textit{clay} objects are treated as threat to detect. It is noteworthy that the multi-class classification results can be directly applied to varying tasks of different threat definitions (e.g., only one of the three materials is defined as threat), while the binary classification cannot unless the classifier is re-trained hence limiting its adaptability.

As described in Section \ref{sect:data}, we divide the public dataset into two subsets and use one of them for training and the other for testing. During training, we use ground truth voxel annotations to extract the volumes of objects in the training data, and extract the normalised histogram features from the object volumes for the classifier training. Again, we use probability of detection and probability of false alarm as the evaluation metrics for classification performance. Good classification will lead to a reduced probability of false alarm without affecting the probability of detection too much when comparing to those before applying the classification. 

The experimental results are shown in Table \ref{table:cls1} including the threat detection performance without classification, with the binary classification and with the multi-class classification. The use of binary classification significantly reduces the overall probability of false alarm from $0.56$ down to $0.29$, and keeps high probabilities of detection at the same time.  When the multi-class classification is used, the probabilities of false alarm can be further reduced to $0.24$ whilst the probabilities of detection only decrease slightly. These results validate the effectiveness of our multi-class classifier which is essential for the adaptability of ATR. 

\begin{table*}[!t]
	\centering
	{
		\caption[]{Material classification performance on the public data.\\
		}
		\label{table:cls1}
		\begin{lrbox}{\tablebox}
			\begin{tabular}{c|c|ccc|cc|c|c}
				\hline
				\multirow{2}{*}{\textbf{Method}}&\multirow{2}{*}{\textbf{Test Set}} &  \multicolumn{6}{c|}{\textbf{PD}} & \textbf{PFA}\\ \cline{3-9}
				& &\textbf{saline} & \textbf{rubber}   & \textbf{clay}  & \textbf{bulk} & \textbf{sheet} &\textbf{overall} & \textbf{overall} \\
				
				\hline
				w/o classification & all& 0.90 & 0.96 & 0.97  & 0.93 & 0.95 & 0.94 & 0.56 \\
				\hline
				\multirow{3}{*}{binary classification}&odd & 0.90& 0.99 &0.98 &0.94 & 0.99 & 0.96 & 0.28 \\
				&even& 0.90 & 0.93 & 0.93 & 0.92&0.91&0.92 &0.31 \\
				&all & 0.90 & 0.96 & 0.96 & 0.93 & 0.95 & 0.94 & 0.29 \\
				\hline
				\multirow{3}{*}{multi-class classification}&odd& 0.86 &0.98 & 0.98 & 0.92 & 0.97 & 0.94 & 0.22 \\
				&even& 0.87 & 0.92 & 0.93 & 0.91 & 0.89 & 0.91 & 0.26\\
				&all & 0.87 & 0.95 & 0.96 & 0.92 & 0.93 & 0.92 & 0.24\\
				\hline
			\end{tabular}
		\end{lrbox}
		\scalebox{0.8}{\usebox{\tablebox}}
	}
\end{table*}

\subsection{Evaluation of Adaptation}
\label{sect:evalAdapt}
We conduct extensive experiments to evaluate the adaptability of the proposed approach in terms of adaptability metrics defined by ALERT \cite{to7} to simulate varying practical scenarios. All experiments in this section are carried out by employing the proposed segmentation algorithm in Section \ref{sect:segmentation} followed by the multi-class classification. Again, the performance is evaluated in terms of probability of detection and probability of false alarm. Apart from the experiment on unknown materials which is trained on the public dataset and tested on sequestered dataset, all other experiments in the following sub-sections follow the same training/testing scheme as before, i.e., training on one subset (odd/even set) and test on the other (even/odd set) of the public dataset.

\subsubsection{Varying Probability of Detection Requirements}
An ideal detection algorithm is expected to have 100\% probability of detection and 0\% probability of false alarm. Improving the probability of detection usually causes an increase of probability of false alarm, and reducing the probability of false alarm leads to a lower probability of detection.
\begin{table}[!t]
	\centering
	{\caption[]{Results of the adaptation to varying probability of detection requirements.\\}
		\label{table:varyingPD}
		\begin{lrbox}{\tablebox}
			\begin{tabular}{c|c|c|cc}
				\hline
				\textbf{Threat Definition}&\textbf{Required PD} &\textbf{Test Set} &  \textbf{PD}& \textbf{PFA}\\ 
				\hline
				\multirow{15}{*}{\makecell{saline\\(minMass:137, \\Rho:1050-1215)}}&\multirow{3}{*}{0.70}&odd & 0.72 & 0.08 \\
				& &even& 0.72 & 0.08\\
				& &all & 0.72 & 0.08 \\
				
				\cline{2-5}
				&\multirow{3}{*}{0.80}&odd & 0.76 & 0.10 \\
				& &even& 0.80 & 0.11\\
				& &all & 0.78 & 0.10 \\
				\cline{2-5}
				&\multirow{3}{*}{0.85}&odd & 0.81 & 0.12 \\
				& &even& 0.86 & 0.13\\
				& &all & 0.83 & 0.12 \\
				\cline{2-5}
				&\multirow{3}{*}{0.90}&odd & 0.83 & 0.16 \\
				& &even& 0.87 & 0.19\\
				& &all & 0.85 & 0.17 \\
				\cline{2-5}
				&\multirow{3}{*}{0.95}&odd & 0.85 & 0.17 \\
				& &even& 0.87 & 0.21\\
				& &all & 0.86 & 0.19 \\
				\hline
			\end{tabular}
		\end{lrbox}
		\scalebox{0.8}{\usebox{\tablebox}}
	}
\end{table}

Firstly, we try to find optimal values of $\alpha$ in Eq. (\ref{eq:alpha}) for each known material. To this end, the values of $\alpha$ are selected from a candidate set of \{1.0, 0.9, 0.8, 0.7\} exhaustively. Starting from $\alpha=1$, by decreasing the value of $\alpha$, we may achieve a higher probability of detection and a higher probability of false alarm. The optimal value of $\alpha$ is selected when the probability of detection comes to its maximum and an even smaller $\alpha$ will only lead to the increase of false alarms but no further improvement of detection. In our experiments, we find the optimal $\alpha$ to be $0.9$ for \textit{saline} and $0.8$ for \textit{rubber} and \textit{clay}.  

Our AATR approach enables the trade-off of probability of detection and probability of false alarm by manipulating the output probabilities of different threats and non-threats (see Section \ref{sect:adaptation}). To learn the relation (i.e. $f_c(t_{PD})$ in Eq.(\ref{eq:adjustProb})) between probability of detection and the offset value for each material from training data, we set the offset value in Eq.(\ref{eq:adjustProb}) as $\{-0.5, -0.4, ..., 0.4, 0.5\}$ respectively to get the corresponding probabilities of detection, assuming only one type of material is the threat in each experiment. Based on the pairs of probability of detection and offset values, we can have the offset as a function of the target probability of detection, i.e., $f_c(t_{PD})$ for each material class $c$. In the testing phase, the functions are used to calculate the offset values for the given required probabilities of detection to enable our AATR adaptive to the requirements. We use odd/even set to learn the functions and test them on the even/odd set. 

\begin{figure*}[h]	
	\centering
	\includegraphics[width=\textwidth]{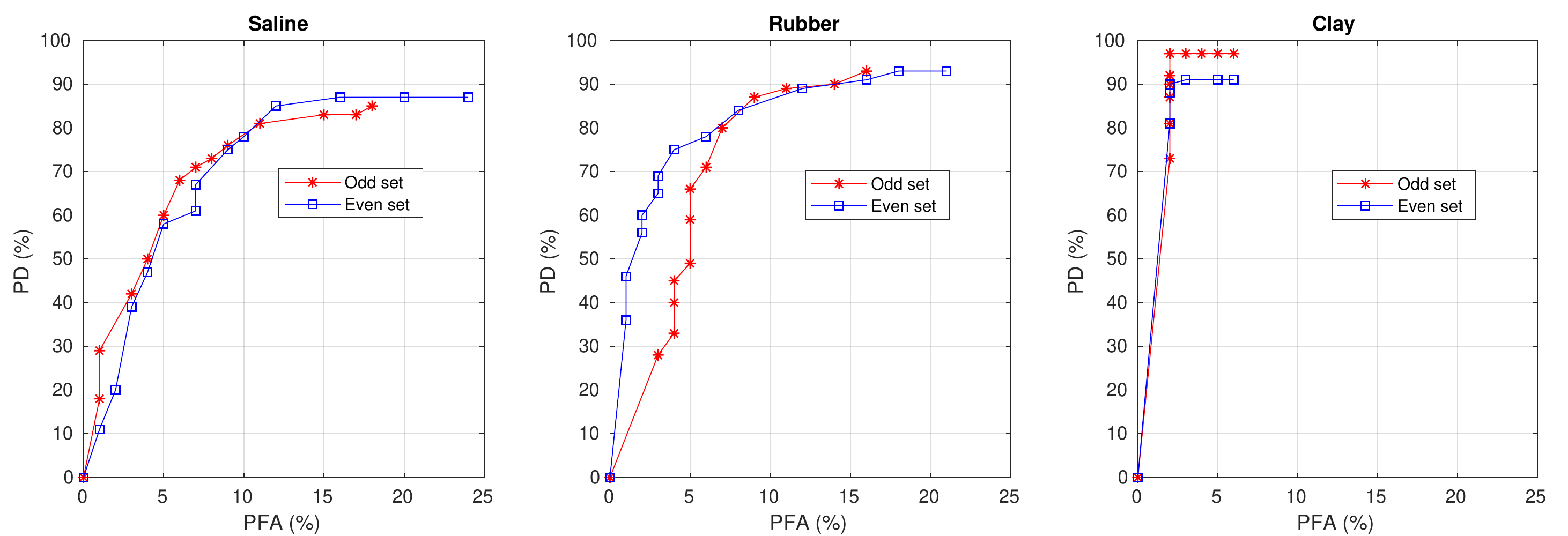}
	\caption{Probability of detection and probability of false alarm achieved by our approach for three different materials.}
	\label{fig:pdpfa}
\end{figure*}

Figure \ref{fig:pdpfa} shows the relations between probability of detection and probability of false alarm for each type of material. Our algorithm achieves the highest probability of detection of 85\% and 87\% on the odd and even set with the probability of false alarm of 18\% and 16\% respectively, when \textit{saline} is the only threat material. The highest probability of detection our algorithm can achieve for \textit{rubber} is 93\% for both odd and even set with the probability of false alarm of 16\% and 18\% respectively. For the \textit{clay} material, the highest probability of detection of 97\% and 91\% can be achieved with a very low probability of false alarm of 2\% on the odd and even set respectively. For both \textit{saline} and \textit{rubber}, we can see from Figure \ref{fig:pdpfa} that the probability of detection and probability of false alarm gradually change with different offset values in Eq. (\ref{eq:adjustProb}). The behaviour of \textit{clay} is quite different in that the change of offset in Eq.(\ref{eq:adjustProb}) does not affect the probability of detection and probability of false alarm very much. The reason is \textit{clay} objects are separated from others with a large margin, thus it is easier for the classifier to make the decision with high confidence, i.e., the probability is close to one for a positive object and zero for a negative one. Therefore changing the probabilities slightly does not affect the classification results very much.

\begin{figure*}[h]	
	\centering
	\includegraphics[width=\textwidth]{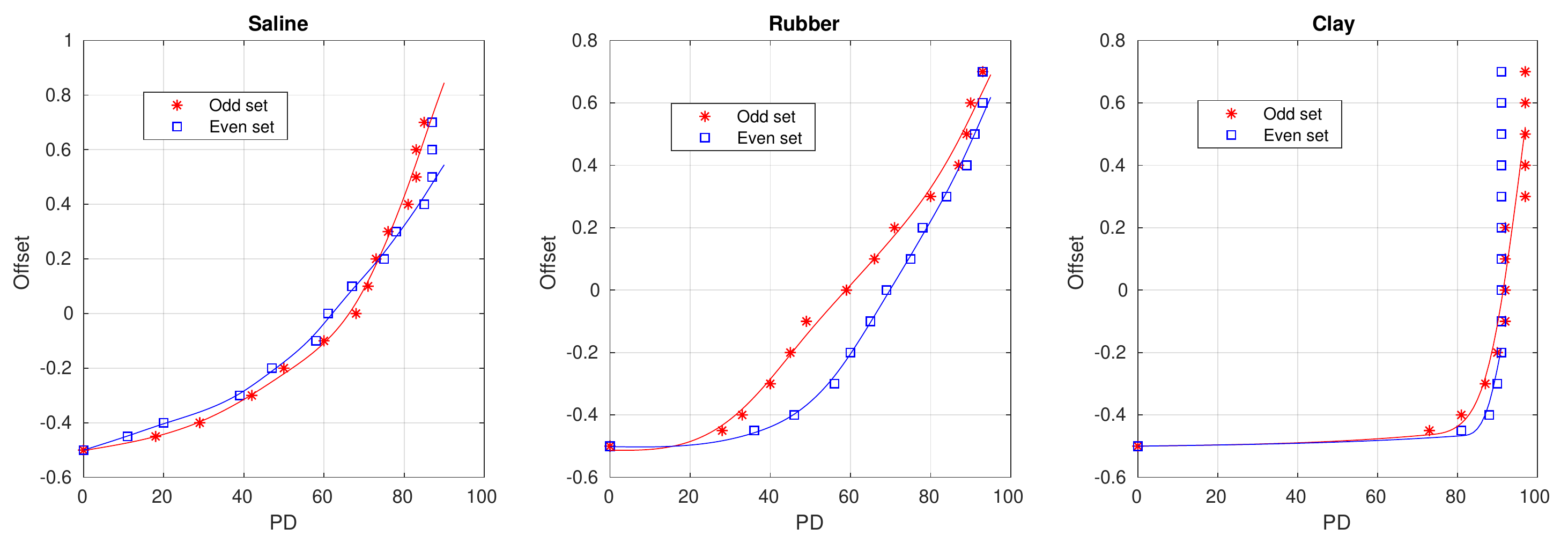}
	\caption{The relations between offset and probability of detection for three different materials.}
	\label{fig:offset}
\end{figure*}

Figure \ref{fig:offset} shows the relations between offset values and the probability of detection for different materials. In each sub-graph, we have pairs of probability of detection and offset values obtained from the experiments on training data (i.e. odd set in red and even set in blue), based on which we can infer the offset functions of target probability of detection for different materials. In our experiments, we use the smoothing spline approach to fit the data points and get the offset functions. The offset function learned from the odd set will be used for evaluation experiments on the even set, and vice versa.

To evaluate the adaptability of our approach to varying probability of detection, we set the target probability of detection for \textit{saline} as \{0.7,  0.8, 0.85, 0.9, 0.95\} respectively, and estimate the offset value for output probability manipulation in Eq.(\ref{eq:adjustProb}) using the learned offset functions. The experimental results are shown in Table \ref{table:varyingPD}. We can see our approach achieves 0.72 and 0.78 probability of detection when the requirements are 0.7 and 0.8 respectively. When the requirement increases to 0.9 and 0.95, our approach achieves  0.85 and 0.86 respectively which has been the best performance of this approach. In summary, our approach is able to adapt itself to satisfy varying probability of detection requirements unless the required probability of detection is beyond the upper performance bound of the approach.

\subsubsection{Varying Threat Materials}
This experiment aims to evaluate how our approach adapts to varying threat materials. We simulate this scenario by defining the threats in different ways which are shown in the first column of Table \ref{table:varyingOOI}. Each definition specifies the materials that could potentially be threats as well as the requirements of physical properties (i.e., minimum mass (\textit{minMass}) and density ($Rho$) range) that make an object be a threat for each material. In all cases, the required probability of detection is set as 0.9 and the learned offset functions are used to estimate the offset values in Eq.(\ref{eq:adjustProb}). There are totally six different definitions of threats in Table \ref{table:varyingOOI} to simulate six scenarios of threat recognition. 

\begin{table*}[!t]
	\centering
	{\caption[]{Results of the adaptation to varying threat material signatures.\\}
		\label{table:varyingOOI}
		\begin{lrbox}{\tablebox}
			\begin{tabular}{ccc|c|ccc|cc|c|c}
				\hline
				\multicolumn{3}{c|}{\textbf{Threat Definition}}&\multirow{2}{*}{\textbf{Test Set}} &  \multicolumn{6}{c|}{\textbf{PD}} & \textbf{PFA}\\ \cline{1-3}\cline{5-11}
				\textbf{Material}&\textbf{minMass}&\textbf{Rho range}& &\textbf{saline} & \textbf{rubber}   & \textbf{clay}  & \textbf{bulk} & \textbf{sheet} &\textbf{overall} & \textbf{overall} \\
				
				\hline
				\multirow{3}{*}{\makecell{saline\\ rubber\\ clay}}& \multirow{3}{*}{\makecell{250\\250\\250}}& \multirow{3}{*}{\makecell{1000-2000\\1000-2000\\1000-2000}}&odd & 0.87 & 0.99 & 0.98 & 0.93 & 0.99 & 0.95 & 0.22\\
				\cline{4-11}
				&&&even& 0.90 & 0.97 & 0.94 & 0.93 & 0.96 & 0.94 & 0.26\\
				\cline{4-11}
				&&&all & 0.88 & 0.98 & 0.96 & 0.93 & 0.97 & 0.94 & 0.24 \\
				\hline \hline
				\multirow{3}{*}{\makecell{saline\\ rubber\\ clay}}&\multirow{3}{*}{\makecell{137\\ 180\\ 80}}&\multirow{3}{*}{\makecell{1050-1215\\ 1170-1290\\ 1530-1715}}&odd&0.86 & 0.98 & 0.92 & 0.90 & 0.97 & 0.92 & 0.21\\
				\cline{4-11}
				&&&even& 0.87 & 0.92 & 0.91 & 0.91 & 0.88 & 0.90 & 0.25 \\
				\cline{4-11}
				&&&all & 0.87 & 0.95 & 0.92 & 0.90 & 0.93 & 0.91 & 0.23\\
				\hline \hline
				\multirow{3}{*}{\makecell{saline\\ rubber}}&\multirow{3}{*}{\makecell{137\\ 180}}&\multirow{3}{*}{\makecell{1050-1215\\1170-1290}}&odd& 0.86 & 0.98 & - & 0.89 & 0.97 & 0.92 & 0.18 \\
				\cline{4-11}
				&&&even& 0.87 & 0.92 & - & 0.90 & 0.89 & 0.90 & 0.20 \\
				\cline{4-11}
				&&&all & 0.87 & 0.95 & - & 0.95 & 0.89 & 0.93 & 0.19\\
				\hline\hline
				\multirow{3}{*}{\makecell{saline}}&\multirow{3}{*}{\makecell{137}}&\multirow{3}{*}{\makecell{1050-1215}}&odd& 0.83 & - & - & 0.83 & 1.00 & 0.83 & 0.16\\
				\cline{4-11}
				&&&even& 0.87 & - & - & 0.87 & 1.00 & 0.87 & 0.19\\
				\cline{4-11}
				&&&all & 0.85 & - & - & 0.85 & 1.00 & 0.85 & 0.17\\
				\hline\hline
				\multirow{3}{*}{\makecell{rubber}}&\multirow{3}{*}{\makecell{180}}&\multirow{3}{*}{\makecell{1170-1290}}&odd& - & 0.89 & - & 100 & 0.85 & 0.89 & 0.11 \\
				\cline{4-11}
				&&&even& - & 0.91 & - & 1.00 & 0.88 & 0.91 & 0.15\\
				\cline{4-11}
				&&&all & - & 0.90 & - & 1.00 & 0.86 & 0.90 & 0.13\\
				\hline\hline
				\multirow{3}{*}{\makecell{clay}}&\multirow{3}{*}{\makecell{80}}&\multirow{3}{*}{\makecell{1530-1715}}&odd& - & - & 0.87 & 0.85 & 1.00 & 0.87 & 0.02\\
				\cline{4-11}
				&&&even& - & - & 0.90 & 0.92 & 0.80 & 0.90 & 0.02\\
				\cline{4-11}
				&&&all & - & - & 0.88 & 0.88 & 0.89 & 0.88 & 0.02\\
				\hline
			\end{tabular}
		\end{lrbox}
		\scalebox{0.8}{\usebox{\tablebox}}
	}
\end{table*}

In the first two scenarios, three materials are defined as threats. The differences between these two scenarios lie in the specified minimum masses (\textit{minMass}) and density ($Rho$) ranges. The results in Table \ref{table:varyingOOI} show that our approach performs well in both scenarios. The performance for the second scenario is slightly worse than that for the first scenario in terms of probability of detection (PD). The reason is that smaller objects are required to be detected in the second scenario, which poses more challenges. On the other hand, the probabilities of false alarm achieved in the second scenario are reduced due to the additional restrictions of density ranges in this scenario so that many non-threat signatures can be correctly excluded from the final detection results.

In the third scenario, two materials of ``\textit{saline}" and ``\textit{rubber}"  are defined as threats with the material of ``\textit{clay}" defined as non-threat. Benefiting from the material classification employed in our approach, the threat materials can be detected with a comparable probability of detection to that in the second scenario and the probabilities of false alarm are also significantly reduced since non-threat object signatures with densities close to that of ``\textit{clay}" can be easily excluded.

Finally, in the last three scenarios only one material is defined as threat. Significant drops of probability of false alarm can be observed from Table \ref{table:varyingOOI} when comparing to that in the second scenario where all three materials are defined as threats. The probabilities of detection also decrease by considerable margins. This is due to the overlapping intensity distributions of different materials and classification errors. For instance, in the second scenario, a true ``\textit{saline}" signature can still have a chance to be reserved as a positive detection even if it is misclassified into ``\textit{rubber}" or ``\textit{saline}" since all three materials are defined as threats, in the last three scenarios, however, it will definitely be mistakenly missed once it is misclassified hence leading to lower probability of detection.

Overall, the experimental results in Table \ref{table:varyingOOI} clearly indicate our proposed AATR approach has the adaptability to varying material requirements.

\subsubsection{Varying Minimum Mass/Thickness}
This experiment aims to evaluate how our approach adapts to varying physical properties (i.e. mass and thickness) of the threats.
The proposed multi-scale segmentation algorithm (see Section \ref{sect:shapesplit}) has the capability of detecting objects of varying scales, thus enabling our AATR adaptive to threats of varying mass/thickness. 

To investigate how the approach performs in terms of varying masses in detail, we define three threat definitions with varying masses. We consider ``\textit{saline}" as the only threat material and set the mass range as $400-Inf$, $300-400$ and $100-300$ as the requirements for being a threat respectively. The experimental results are shown in Table \ref{table:varyingMass}. Our approach is able to achieve the probability of detection of 0.94 and 0.93 when the threats have masses within the range of $400-Inf$ and $300-400$, respectively. Apparently, threats with lower masses are more difficult to detect, but our approach can still achieve reasonable good performance with a probability of detection of 0.80 for the threats within the mass range of $100-300$.

A similar investigation is also made on how our approach performs in terms of varying thickness. In this experiment, we consider ``\textit{rubber}" as the only threat material since there are a reasonable amount of \textit{rubber} sheets in the dataset. The thickness ranges of $10-Inf$, $6.5-10$ and $0-6.5$ are selected as the requirements for being a threat respectively. Experimental results are shown in Table \ref{table:varyingThickness}. Our approach is able to achieve the probability of detection of 0.90 and 0.89 when the threats have thickness within the range of $10-Inf$ and $6.5-10$, respectively. Again, thinner threats within the thickness range of $0-6.5$ are more difficult to detect in which case our approach achieves reasonable well performance with a probability of detection of 0.84.

\begin{table}[h]
	\centering
	{\caption[]{Results of the adaptation to varying Mass.\\}
		\label{table:varyingMass}
		\begin{lrbox}{\tablebox}
			\begin{tabular}{ccc|c|c|cc}
				\hline
				\multicolumn{3}{c|}{\textbf{Threat Definition}}&\multirow{2}{*}{\textbf{Test Set}} & \multirow{2}{*}{ \textbf{PD}}& \multirow{2}{*}{\textbf{PFA}}\\ 
				\cline{1-3}
				\textbf{Material}&\textbf{Rho range}&\textbf{Mass range}& &&\\
				\hline
				\multirow{3}{*}{saline} &\multirow{3}{*}{1050-1215}&\multirow{3}{*}{400-Inf} & odd & 0.85 & 0.16 \\
				\cline{4-6}
				&&&even& 1.00 & 0.19\\
				\cline{4-6}
				&&&all & 0.94 & 0.17 \\
				\hline				
				\hline
				\multirow{3}{*}{saline}&\multirow{3}{*}{1050-1215}&\multirow{3}{*}{300-400}&odd & 0.92 & 0.16 \\
				\cline{4-6}
				&&&even& 0.94 & 0.19\\
				\cline{4-6}
				&&&all & 0.93 & 0.17 \\
				\hline
				\hline
				\multirow{3}{*}{saline}&\multirow{3}{*}{1050-1215}&\multirow{3}{*}{100-300}&odd & 0.82 & 0.16 \\
				\cline{4-6}
				&&&even& 0.77 & 0.19\\
				\cline{4-6}
				&&&all & 0.80 & 0.17 \\
				
				\hline
			\end{tabular}
		\end{lrbox}
		\scalebox{0.8}{\usebox{\tablebox}}
	}
\end{table}

\begin{table}[h]
	\centering
	{\caption[]{Results of the adaptation to varying Thickness.\\}
		\label{table:varyingThickness}
		\begin{lrbox}{\tablebox}
			\begin{tabular}{ccc|c|c|cc}
				\hline
				\multicolumn{3}{c|}{\textbf{Threat Definition}}&\multirow{2}{*}{\textbf{Test Set}} & \multirow{2}{*}{ \textbf{PD}}& \multirow{2}{*}{\textbf{PFA}}\\ 
				\cline{1-3}
				\textbf{Material}&\textbf{Rho range}&\textbf{Thickness}& &&\\
				\hline
				\multirow{3}{*}{rubber}&\multirow{3}{*}{1170-1290}&\multirow{3}{*}{10-Inf}&odd & 0.83 & 0.11 \\
				\cline{4-6}
				&&&even& 1.00 & 0.15\\
				\cline{4-6}
				&&&all & 0.90 & 0.13 \\
				\hline
				\hline
				\multirow{3}{*}{rubber}&\multirow{3}{*}{1170-1290}&\multirow{3}{*}{6.5-10}&odd & 0.80 & 0.11 \\
				\cline{4-6}
				&&&even& 1.00 & 0.15\\
				\cline{4-6}
				&&&all & 0.89 & 0.13 \\
				\hline
				\hline
				\multirow{3}{*}{rubber}&\multirow{3}{*}{1170-1290}&\multirow{3}{*}{0-6.5}&odd & 0.76 & 0.11 \\
				\cline{4-6}
				&&&even& 0.96 & 0.15\\
				\cline{4-6}
				&&&all & 0.84 & 0.13 \\
				
				\hline
			\end{tabular}
		\end{lrbox}
		\scalebox{0.8}{\usebox{\tablebox}}
	}
\end{table}

\subsubsection{Unknown Threat Materials}
One important requirement of the next generation AATR algorithms for baggage screening is to improve the detection performance of unknown threat materials which are not seen during training phase. The proposed approach in this paper has the ability to recognize unknown threat material signatures due to its unsupervised segmentation and flexible classification algorithms. In this experiment, we test our algorithms on the sequestered dataset which consists of four different types of unknown materials named \textit{m1}, \textit{m2}, \textit{m3} and \textit{m4}. We train our approach on the public dataset and test it on sequestered dataset in terms of the detection of unknown material signatures. The experimental results are shown in Table \ref{table:unknown}. Our AATR achieves high probability of detection of 76\%, 100\%, 92\%, 100\% for four types of unknown materials with the probability of false alarm of 12\%, 46\%, 15\% and 11\% respectively. It can be seen the performance on \textit{m3} and \textit{m4} are better than that on \textit{m1} and \textit{m2}. The probability of detection for m1 is 76\% which is lower than average, probably due to the small scales of those object signatures in the data. On the other hand, the probability of false alarm for \textit{m2} is as high as 0.46 although the probability of detection is perfect. One reasonable explanation is that many non-threat objects have overlapped intensity distribution with this material, thus making it difficult to distinguish between these threat and non-threat signatures. 

\begin{table}[h]
	\centering
	{\caption[]{Results of the adaptation to unknown materials.\\}
		\label{table:unknown}
		\begin{lrbox}{\tablebox}
			\begin{tabular}{ccc|cc}
				\hline
				\multicolumn{3}{c|}{\textbf{Threat Definition}}& \multirow{2}{*}{ \textbf{PD}}& \multirow{2}{*}{\textbf{PFA}}\\ 
				\cline{1-3}
				\textbf{Material}&\textbf{minMass}&\textbf{Rho range}& &\\
				\hline
				m1&42&380-525& 0.76 & 0.12 \\
				m2&67&770-810& 1.00 & 0.46\\
				m3&174&1300-1375& 0.92 & 0.15\\
				m4&183&1350-1430& 1.00 & 0.11\\
				
				\hline
			\end{tabular}
		\end{lrbox}
		\scalebox{0.8}{\usebox{\tablebox}}
	}
\end{table}

We also compare the performance of our approach with those of \cite{manerikar2018adaptive} and \cite{paglieroni2018consensus} under the same setting. \cite{manerikar2018adaptive} achieved the probabilities of detection of 27\%, 57\%, 38\% and 55\% for \textit{m1-m4} respectively \cite{manerikar-slides} which are significantly lower than ours, while \cite{paglieroni2018consensus} achieved the probabilities of detection of 94\%, 84\%, 85\% and 80\% \cite{paglieroni-slides} which are lower than ours for three out of four materials but also with lower probabilities of false alarms than ours.  
Overall, it is indicated that our proposed AATR is promising in the adaptability to unknown materials which are not available in training data.

\section{Discussion}
\label{sect:discussion}
In this section, we discuss the potential issues underlying the proposed approach.
One issue within X-ray CT image segmentation and classification are the metal artefacts caused by metal material in the baggage \cite{mouton2018relevance}. In our experiment, we focus more on the adaptability to varying requirements than the segmentation performance. Since the data used in our experiments were captured by a medical CT machine which suffers less from the metal artefacts, there is a risk of performance drop when the segmentation algorithm is applied to baggage data with significant artefacts. A potential solution to addressing this issue is to apply the metal artefact reduction algorithms \cite{naidu2004method, mouton2013experimental} as a pre-process, though this could be marginal and yet require considerable additional computation \cite{mouton2018relevance}.

The proposed approach involves some parameters which are empirically selected based on the training data, leading to a risk of over-fitting to the data we currently have. 
In the segmentation algorithm, there is no training process but some very important parameters whose values are empirically selected. First, the parameter values of the structuring element used in the multi-scale morphological opening operation are set as 2, 3 and 8 in our experiments. These values work under the assumption that all the threat signatures should have volumes bigger than the smallest-scale structuring element so that they cannot be removed when applying the erosion operation. If the material of interest has smaller volumes, poorer performance can be expected. The other parameter in the segmentation algorithm, voxel number threshold, has a similar effect on the performance. Currently, the value of this parameter is set as $80\times 10^3$ empirically, assuming that any segmented object having less than this volume will be non-threats. The potential over-fitting issues raised by these parameter values could be partly eliminated with prior knowledge of the task at hand. For example, if we know, in prior, the minimum volume of threat materials, proper values of these parameters can be decided.

In the classification algorithm, we need to train a four-class SVM classifier to recognize the material of a given segmented object. The classification accuracy is important for a low probability of false alarm by distinguishing non-threats from threats. It is also important for a high probability of detection by not mistakenly classifying a threat segment to be the non-threat. Our SVM classifier is trained on the ground truth threat objects in the training data plus synthesized features for unknown materials based on their density ranges. We use only the normalised histogram as the features for classification. As we can see from Figure \ref{fig:intensityDist}, there are overlapping of features between different materials, hence learning a large margin between classes would be favourable to fight off the over-fitting issue. 

Evaluations in this work are based on a relatively small dataset due to the challenge of collecting and annotating a large-scale CT baggage dataset. As a proof of concept, the proposed approach can achieve superior performance compared with other works \cite{atrAlert,manerikar2018adaptive,paglieroni2018consensus} using the same dataset. However, more cluttered CT images and more types of threat materials will be involved in the real-world applications. To address these issues we will consider improving the 3D segmentation algorithms and employing more powerful material classification methods within the proposed AATR framework in our future work. 

\section{Conclusion}
\label{sect:conclusion}
In this paper, we propose a solution to the adaptive automatic threat (explosive materials) recognition (AATR) problem. We propose an unsupervised segmentation algorithm for 3D CT images which works well on objects of different materials. Based on the segmentation results, our approach is able to adapt to the requirements of varying threat materials, varying probability of detection and the capability of detecting new materials, without the need of re-training on the labelled data.

In our future work, the classification performance could be improved by using more features than only the normalised histogram or using dual-/multi-energy  CT images. In our current approach, we have not explicitly handled the CT image noises such as metal artefacts which have been reported affecting the segmentation \cite{martin2015learning}. To enhance the adaptability to new materials, the techniques of zero-shot learning \cite{wang2017zero} and few-shot learning \cite{snell2017prototypical} can also be employed in future work.

\bibliographystyle{plain}
\bibliography{ref}
\end{document}